\documentclass[10pt,twocolumn,letterpaper]{article}

\usepackage{cvpr}              

\usepackage[dvipsnames]{xcolor}

\definecolor{cvprblue}{rgb}{0.21,0.49,0.74}
\usepackage[pagebackref,breaklinks,colorlinks,citecolor=cvprblue]{hyperref}

\usepackage{graphicx}
\usepackage{amsmath}
\usepackage{amssymb}
\usepackage{booktabs}
\usepackage{cuted}
\usepackage{multirow}
\usepackage{multicol}

\def\paperID{395} 
\def\confName{3DV\xspace}
\def\confYear{2025\xspace}

\title{DressRecon: Freeform 4D Human Reconstruction from Monocular Video}

\author{{Jeff Tan, Donglai Xiang, Shubham Tulsiani, Deva Ramanan, Gengshan Yang}\thanks{Corresponding author: \texttt{jefftan@andrew.cmu.edu}}\\
Carnegie Mellon University, USA\\
}
\begin{document}
\maketitle
\begin{strip}
    \centering
    \includegraphics[width=0.35\linewidth, trim={0.5cm 1cm 13cm 5.5cm}, clip, angle=270]{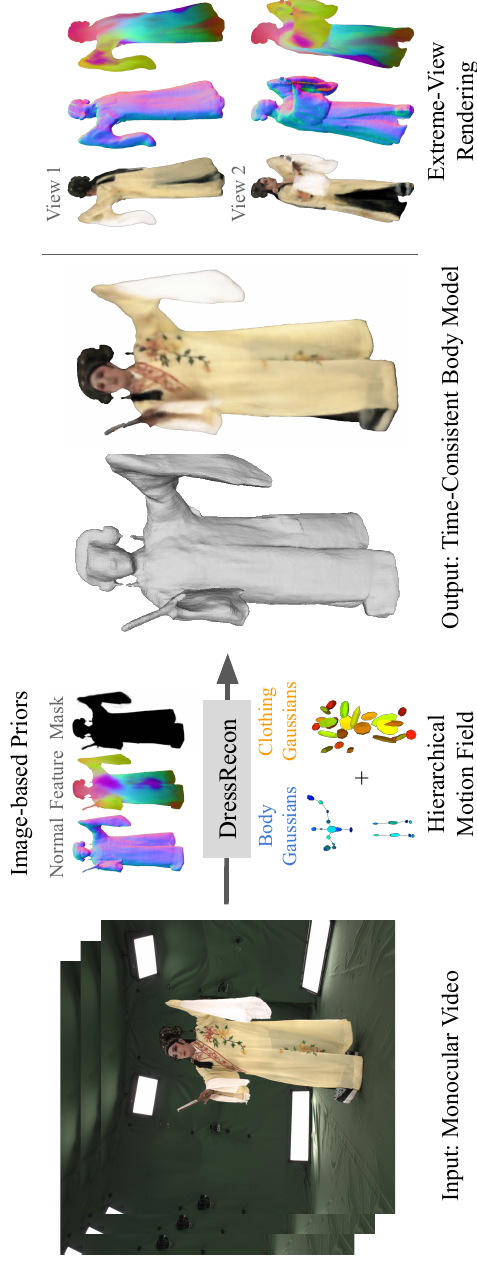}
    \captionof{figure}{
    Given an input video of a human, DressRecon reconstructs a time-consistent 4D body model, including shape, appearance, time-varying body articulations, as well as deformation of extremely loose clothing or accessory objects. We propose a hierarchical bag-of-bones deformation model that allows body and clothing motion to be separated. We leverage image-based priors such as human body pose, surface normals, and optical flow to make optimization more tractable. The resulting neural fields can be extracted into time-consistent meshes, or further optimized as explicit 3D Gaussians for high-fidelity interactive rendering.
    }
    \label{fig:teaser}
\end{strip}

\begin{abstract}
We present a method to reconstruct time-consistent human body models from monocular videos, focusing on extremely loose clothing or handheld object interactions. Prior work in human reconstruction is either limited to tight clothing with no object interactions, or requires calibrated multi-view captures or personalized template scans which are costly to collect at scale. Our key insight for high-quality yet flexible reconstruction is the careful combination of generic human priors about articulated body shape (learned from large-scale training data) with video-specific articulated ``bag-of-bones" deformation (fit to a single video via test-time optimization). We accomplish this by learning a neural implicit model that disentangles body versus clothing deformations as separate motion model layers. To capture subtle geometry of clothing, we leverage image-based priors such as human body pose, surface normals, and optical flow during optimization. The resulting neural fields can be extracted into time-consistent meshes, or further optimized as explicit 3D Gaussians for high-fidelity interactive rendering. On datasets with highly challenging clothing deformations and object interactions, DressRecon yields higher-fidelity 3D reconstructions than prior art. Project page: \href{https://jefftan969.github.io/dressrecon/}{https://jefftan969.github.io/dressrecon/}
\end{abstract}

\section{Introduction}
\label{sec:intro}

We aim to reconstruct animatable dynamic human avatars from videos of people wearing loose clothing or interacting with objects, such as in-the-wild {\em monocular} videos recorded on a phone or from the Internet. High-quality reconstructions in this setting traditionally require calibrated multi-view captures \cite{xu20234k4d, luiten2023dynamic}, which are costly to obtain.

From only a single viewpoint, recovering freely-deforming humans with arbitrary topology is highly under-constrained, and thus prior works often rely on domain-specific constraints which struggle to support loose clothing. Template-based human reconstruction \cite{habermann2020deepcap, habermann2018livecap, xu2017monoperfcap} requires personalized scanned templates, which works well for a single instance but cannot reconstruct unseen clothing and body shapes and clothing. Methods that regress 3D surfaces from a single image \cite{xiu2023econ, xiu2023icon} can produce high-quality geometry at observed regions, but the results are inconsistent across frames and sometimes fail to produce coherent body shapes. Human-specific methods \cite{guo2023vid2avatar, weng2022humannerf,jiang2022neuman} can achieve high quality on tight clothing, but often use a fixed human skeleton or parametric body template and thus cannot handle extreme deformations outside the body. More broadly, generic methods for humans and animals \cite{yang2021viser, yang2021lasr} can support arbitrary deformations, but often produce lower quality results than human-specific methods.

This paper presents DressRecon, which reconstructs freeform 4D humans with loose clothing and handheld objects from monocular videos. Our key insight is the careful combination of generic human-level priors about articulated body shape (learned from large-scale training data) with video-specific articulated ``bag-of-bones" clothing models (fit to a single video via test-time optimization). We accomplish this by learning a neural implicit model that disentangles body and clothing deformations as separate motion layers. To capture subtle geometry of clothing, we leverage image-based priors such as masks, normals, and body pose during optimization. When the goal is shape reconstruction, we extract time-consistent meshes from the optimized neural fields. Otherwise, to enable high-quality interactive rendering, we propose a refinement stage that converts our implicit neural body into 3D Gaussians while maintaining the motion field design. On datasets with highly challenging clothing and object deformations, DressRecon yields higher-fidelity 3D reconstructions than prior art.

\section{Related Work}
\label{sec:related_work}

\noindent\textbf{Humans from multi-view or depth.}
With sufficient information as input, multi-view methods \cite{debevec2000acquiring,joo2017panoptic,peng2021neural,geng2023learning,li2023animatable,luiten2023dynamic,xiang2021modeling} can reconstruct human shape and appearance of very high fidelity, but the reliance on a dense capture studio limits their applicability at a consumer level. Depth-based methods \cite{yu2017bodyfusion,xiang2023drivable,dou2016fusion4d} follow the seminal DynamicFusion work \cite{newcombe2015dynamicfusion} to integrate human shape from a monocular depth stream into a canonical space with the help of a deformation model. However, their application scenarios are also limited because they require specialized depth sensors.

\noindent\textbf{Monocular human reconstruction.} Monocular RGB-based reconstruction is challenging due to the 3D ambiguity of a monocular input. Early work \cite{bogo2016keep,kanazawa2018end,xiang2019monocular,goel2023humans} aims to reconstruct 3D human keypoints or skeletal poses using a deformable human model \cite{loper2015smpl,joo2018total}. Compared with sparse keypoints, reconstructing dense human surfaces is even more challenging, especially when clothing is considered. Trained on ground truth 3D scans, pixel-aligned implicit functions \cite{saito2019pifu,xiu2023econ,li2020monocular} regress clothed human surfaces from a monocular image, but their output on a video tends to be less temporally coherent. Another line of work aims to reconstruct dynamic human shapes from video input, using a deformable human model \cite{guo2023vid2avatar,weng2022humannerf,jiang2022selfrecon} or pre-scanned personalized templates \cite{xu2017monoperfcap,habermann2020deepcap,jiang2022hifecap} and often achieving significant speedups \cite{jiang2022instantavatar,hu2024gaussianavatar,lei2024gart}. Generic human models (e.g. SMPL) help resolve monocular 3D ambiguity, but without a personalized clothed template, few works can handle dynamic clothing that does not closely follow body motion. HOSNeRF~\cite{liu2023hosnerf} reconstructs objects rigidly attached to the human body (e.g., hand) by introducing new object bones into the human skeleton hierarchy. Our method took a step further and introduces a novel representation that not only leverages human-specific model priors, but also simultaneously enjoys the flexibility to handle loose garments. A concurrent work, ReLoo~\cite{reloo}, also applies a two layer deformation model to account for the motion of loose garments.

\noindent\textbf{Monocular nonrigid 3D reconstruction.} 
Non-rigid structure from motion (NRSfM) methods~\cite{bregler2000recovering} reconstruct non-rigid 3D shapes from 2D point trajectories in a class-agnostic way. However, due to the oversimplified motion model and the difficulties in estimating long-range correspondences~\cite{sand2008particle}, they do not work well for videos with challenging deformations. Recent work applies differentiable rendering to reconstruct articulated objects from videos~\cite{pumarola2020d,yang2021lasr,yang2021viser,wu2021dove} or images~\cite{ye2021shelf,goel2020shape,kanazawa2018learning, wu2023magicpony}. However, they cannot reconstruct challenging body articulations and large deformations beyond the body, due to the lack of a flexible motion representation and sufficient measurement signals. As shown in Tab. \ref{tab:related_work}, we introduce a hierarchical bag-of-bones motion model that is capable of representing the deformation of loose garments and accessories, fitted using rich signals from pretrained vision models such as human body pose, surface normals, and optical flow.

\begin{table}[!t]
    \caption{\textbf{Related work} in monocular 3D body reconstruction. $^{(1)}$Methods based on human body and pose models. $^{(2)}$General methods for humans and animals. Dense: Dense deformation fields. Bob: Bag-of-bones. H: Human body and pose priors. F: Optical flow. N: Surface normal. $\boldsymbol\phi$: Features. Our method combines the best of human-specific and general methods by fitting a flexible motion model initialized from off-the-shelf 3D human poses, using dense image-based priors.}
    \footnotesize
    \centering
    \begin{tabular}{rrccc}
    \toprule
    &Method  &  Motion model & Prior & Input \\
    \midrule
    \multirow{5}{*}{\shortstack{(1)}}
    &ECON~\cite{xiu2023econ}   & N.A.         & H,N  & Image \\
    &NeuMan~\cite{jiang2022neuman} & Skeleton & H & Video \\
    &Vid2Avatar~\cite{guo2023vid2avatar}   & Skeleton  & H  & Video \\
    &SelfRecon~\cite{jiang2022selfrecon} & Skeleton+Dense & H,N & Video \\
    &HumanNeRF~\cite{weng2022humannerf} & Skeleton+Dense & H & Video \\
    \midrule
    \multirow{4}{*}{\shortstack{(2)}} 
    &MagicPony~\cite{wu2023magicpony} & Skeleton & $\boldsymbol\phi$ & Image\\
    &LASR~\cite{yang2021lasr} & Bob   & F    &  Video \\
    &BANMo~\cite{yang2022banmo} & Bob   & F,$\boldsymbol\phi$    &  Video \\
    &RAC~\cite{yang2023rac} & Skeleton+Dense  & F,$\boldsymbol\phi$    &  Video \\
    \midrule
    & DressRecon (Ours) & Hierarchical Bob & H,F,N,$\boldsymbol\phi$ & Video\\
    \bottomrule
\label{tab:related_work}
\end{tabular}
\vspace{-20pt}
\end{table}

\begin{figure*}[!h]
    \centering
    \includegraphics[height=0.9\linewidth, trim={0.5cm 1.5cm 12.5cm 3cm},clip,angle=270]{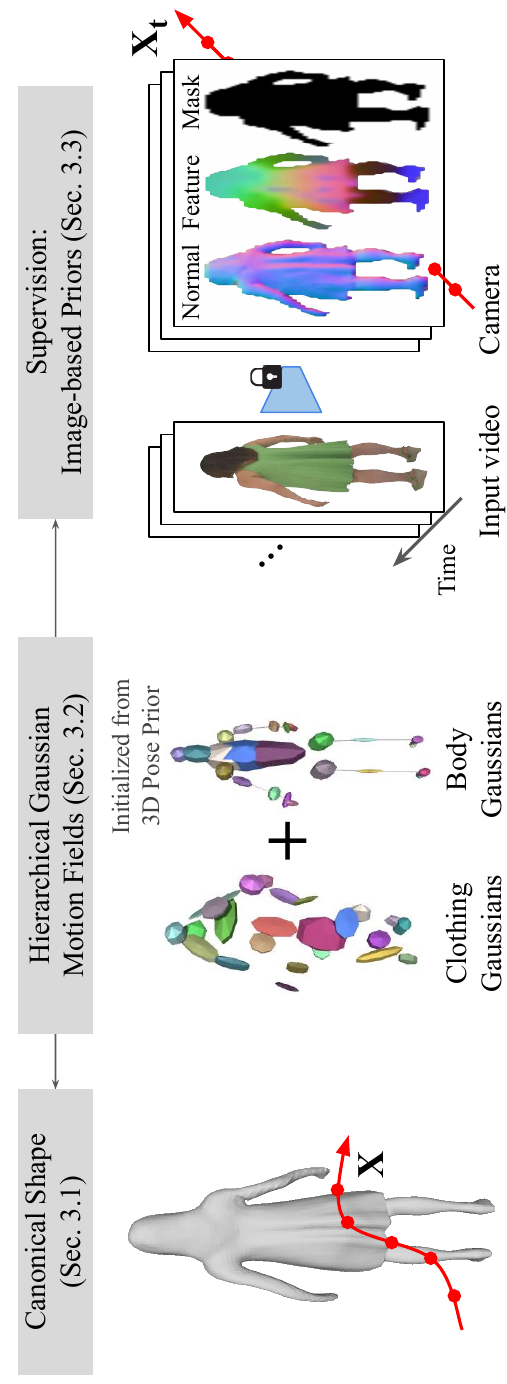}
    \vspace{-5pt}
    \caption{\textbf{Method Overview}: We represent 3D humans in loose clothing as temporally consistent 4D neural fields (Sec.~\ref{sec: preliminary}). Central to our approach is a flexible motion representation that captures fine-grained clothing deformations as well as limb motions, while effectively utilizing domain-specific priors such as 3D human body pose (Sec.~\ref{sec: motion}). We perform video-specific optimization that fits this model to dense image-based priors via differentiable rendering (Sec.~\ref{sec: optimization}). After optimization, our neural implicit surface can be extracted into a time-consistent mesh via marching cubes, or converted into explicit 3D Gaussians for high-fidelity interactive rendering (Sec.~\ref{sec: refinement}).}
\label{fig:method}
\vspace{-10pt}
\end{figure*}

\section{Method}
\label{sec:method}

Our goal is to reconstruct time-varying 3D humans in loose clothing from in-the-wild monocular videos (Fig. \ref{fig:method}). We represent humans with clothing as 4D neural fields and perform per-video optimization with differentiable rendering (Sec.~\ref{sec: preliminary}). Key to our approach is a hierarchical motion model (Sec.~\ref{sec: motion}) capable of representing large limb motions as well as clothing and object deformations. We leverage image-based priors (Sec.~\ref{sec: optimization}) such as body pose, surface normals, and optical flow to make optimization more stable and tractable. The resulting neural fields can be extracted into time-consistent meshes via marching cubes, or converted into explicit 3D Gaussians for high-fidelity interactive rendering (Sec.~\ref{sec: refinement}).

\subsection{Preliminary: Consistent 4D Neural Fields}
\label{sec: preliminary}
To represent a time-varying 3D human, we construct a time-invariant canonical shape that is warped by a time-varying deformation field.

\noindent\textbf{Canonical shape.} We represent the body shape as a neural signed distance field in the canonical space, with the following properties: signed distance $d$, color ${\bf c}$, and universal features ${\boldsymbol\phi}$.  The canonical fields are defined as
\begin{align}
    (d, \boldsymbol{\phi})  &= \textbf{MLP}_\mathrm{SDF}({\bf X}), \label{eq:density}\\
    {\bf c}_t &= \textbf{MLP}_\mathrm{color}({\bf X},\boldsymbol{\omega}_{t}), \label{eq:color}
\end{align}
where {\bf X} is a 3D point in canonical space and ${\boldsymbol \omega}_t$ is a time-varying appearance code specific to each frame.

\noindent\textbf{Space-time warpings}. We represent time-varying motion using continuous 3D deformation fields. A forward deformation field $\mathcal{W}(t)^{+}: {\bf X}\rightarrow {\bf X}_t$ maps a canonical 3D point to time $t$. During volume rendering, rays at time $t$ are traced back to the canonical space using a backward deformation field $\mathcal{W}(t)^{-}: {\bf X}_t\rightarrow {\bf X}$. We use a 3D cycle loss $\mathcal{L}_{\mathrm{cyc}}$ to ensure that $\mathcal{W}(t)^{+}\circ\mathcal{W}(t)^{-}$ is close to identity~\cite{li2021neural,yang2022banmo}.

\noindent\textbf{Volume rendering.} Neural fields can be optimized via differentiable volume rendering~\cite{mildenhall2020nerf}, which renders images and minimizes reconstruction errors (e.g. photometric loss). To provide additional supervision on geometry and motion, we augment the training data with additional signals obtained from off-the-shelf networks, detailed in Sec.~\ref{sec: optimization}.

\begin{figure}[t!]
\includegraphics[width=\linewidth, trim={0cm 22cm 3.5cm 0cm}, clip, angle=0]{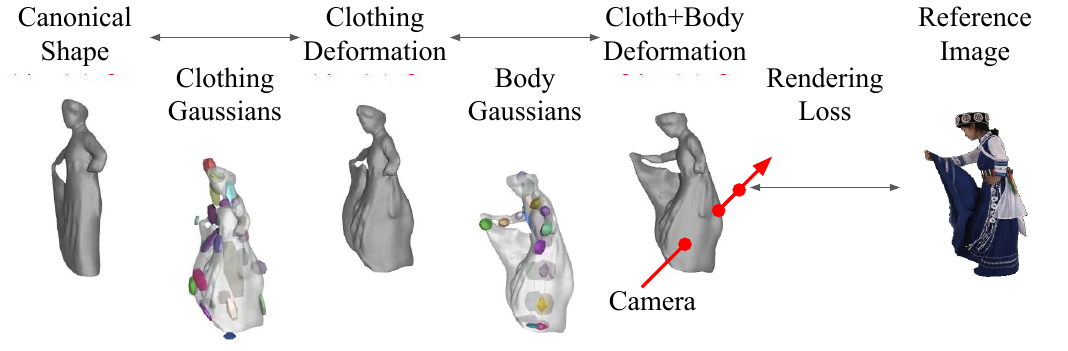}
\caption{
    \textbf{Visualization of two-layer deformation}. The body and clothing deformation layers each contribute separate types of motion. In this sequence, the clothing Gaussians deform the woman's dress to be larger, while the body Gaussians move her right arm forward. During forward warping, we start from the canonical shape (left), and first apply the forward warp described by clothing Gaussians, then the forward warp described by body Gaussians. The same process happens in reverse during backward warping.
}
\label{fig:two_layer_deformation}
\vspace{-10pt}
\end{figure}

\subsection{Hierarchical Gaussian Motion Fields}
\label{sec: motion}
In monocular 4D reconstruction, it is challenging to find a motion representation that is both sufficiently flexible and easy to optimize. Recent methods are either not flexible enough to model dynamic structures outside the body \cite{jiang2022selfrecon}, or struggle to robustly reconstruct dynamic motions at high quality \cite{yang2021lasr}. We introduce hierarchical motion fields to strike a balance between flexibility and robustness.

\noindent\textbf{Bag-of-bones skinning deformation.} 
Our motion model is inspired by deformation graphs and its extension to Gaussian blend skinning models~\cite{sumner2007embedded, bozic2021neural, yang2021lasr}. The idea is to use the motion of $B$ bones (defined as 3D Gaussians, typically $B=25$) to drive the canonical geometry's motion. Each Gaussian maintains a time-varying trajectory of its 3D centers ${\boldsymbol \mu}_t\in\mathbb{R}^{T\times 3}$ and orientations ${\bf V}_t\in\mathbb{R}^{T\times 3}$ over $T$ frames, as well as axis-aligned scales ${\boldsymbol \Lambda}\in\mathbb{R}^{3}$ that are time-invariant. Given the 3D Gaussians, a dense forward deformation field can be computed by blending the $\mathrm{\bf SE}(3)$ transformations of Gaussians with forward skinning weights ${\bf W^+}$. Similarly, a dense backward deformation field is produced by blending with backward skinning weights ${\bf W}^-_t$:
\begin{align}
{\bf X}_t &= \mathcal{W}^{+}({\bf X}, t) = \left(\sum_{b=1}^B {\bf W}^{+,b}{\bf G}^{b}_t\left({\bf G}^b\right)^{-1}\right){\bf X}\label{eq:lbs-fw} \\
{\bf X} &= \mathcal{W}^{-}({\bf X}_t, t)=\left(\sum_{b=1}^B {\bf W}^{-,b}_t{\bf G}^b\left({\bf G}^b_t\right)^{-1}\right){\bf X}_t\label{eq:lbs-bw}
\end{align}

Here ${\bf G}$ and ${\bf G}_t$ are the $\mathrm{\bf SE}(3)$ transformations of the canonical and time $t$ Gaussians, respectively. Forward skinning weights ${\bf W}^+\in\mathbb{R}^b$ are computed using the Mahalanobis distance from ${\bf X}$ to each canonical Gaussian ${\bf G}$. We use a coordinate MLP to refine the weights (similar to \cite{yang2022banmo}), and use a negative softmax such that farther Gaussians are assigned a lower weight. In the same way, backward skinning weights ${\bf W}^-_t\in\mathbb{R}^{T\times b}$ are computed using the Mahalanobis distance from ${\bf X}_t$ to each time $t$ Gaussian ${\bf G}_t$, followed by MLP refinement.

This bag-of-bones representation can represent large non-rigid deformations due to its flexibility, but can be challenging to optimize. For example, most Gaussians can get concentrated in a local region, which limits the ability to deform the other parts of the target. 
Carefully initializing the Gaussians and spatially distributing them during optimization can help avoid such bad local minima. Our key idea is to divide the Gaussians into body and clothing layers, which can be initialized and regularized separately.

\noindent\textbf{Body Gaussians} are intended to represent skeletal motions of the target. With recent advances in human and animal body pose \cite{goel2023humans, nath2019using}, 3D joint locations can be robustly estimated from images and used to initialize the body Gaussian trajectories. This allows body Gaussians to start from a close-to-optimal solution and get locally refined throughout differentiable rendering. The resulting body Gaussians exhibit less temporal jitter than the single-frame predictor, and are better aligned to physical bone locations.

\noindent\textbf{Clothing Gaussians} are intended to represent free-form deformations not explained by body Gaussians, such as cloth deformation and the motion of handheld objects. To encourage that clothing Gaussians only deform structures outside the scope of body Gaussians, we add a regularization term to minimize the impact of clothing Gaussians:
\begin{equation}
\mathcal{L}_{\mathrm{cl}} = \left\|\mathcal{W}^+_{\mathrm{cloth}}({\bf X},t)-{\bf X}\right\|^2
\end{equation}

\noindent\textbf{Compositional two-layer deformation}. The final deformation fields are the composition of body and clothing layer deformations (Fig. \ref{fig:two_layer_deformation}), each with about 25 Gaussian bones. During forward warping we apply the clothing deformation before the body deformation, and during backward warping we perform the reverse:
\begin{align}
\mathcal{W}^{+}(t)=\mathcal{W}^{+}_{\mathrm{body}}( t)\circ\mathcal{W}^{+}_{\mathrm{cloth}}(t)\label{eq:comp-deform-fw}\\
\mathcal{W}^{-}(t)=\mathcal{W}^{-}_{\mathrm{cloth}}(t)\circ\mathcal{W}^{-}_{\mathrm{body}}( t)\label{eq:comp-deform-bw}
\end{align}

We optimize the body and clothing Gaussians jointly. To encourage body and clothing Gaussians to be well-distributed in 3D space, we use a Sinkhorn divergence loss $\mathcal{L}_{\mathrm{sink}}$ \cite{feydy2019interpolating} to match the spatial distribution of Gaussians with the body shape. The Sinkhorn divergence is computed between 1k random points on the canonical rest surface, and 3D points on the Gaussians of each deformation layer. 

With proper initialization and regularization, body and clothing motion can be properly disentangled. In Fig. \ref{fig:two_layer_deformation}, the clothing Gaussians deform the dress while the body Gaussians deform the woman's arm. On the supplementary webpage, we show video examples where body and clothing motion is properly decomposed by two-layer deformation.

\subsection{Optimization with Image-Based Priors}
\label{sec: optimization}
Optimizing time-varying 3D geometry from monocular videos is challenging due to its under-constrained nature. Recent advances in surface normals \cite{eftekhar2021omnidata}, optical flow \cite{yang2019volumetric, teed2020raft}, image features \cite{caron2021emerging, oquab2023dinov2}, and zero-shot segmentation \cite{kirillov2023segment} provide additional interpretations of raw pixel values. This knowledge is not only generic, but also highly correlated with the geometry and motion of the underlying scene, making it suitable for our reconstruction task. We introduce an optimization routine that uses foundational image-based priors as supervision to make the problem tractable.

\noindent\textbf{Surface normals.} Without multi-view inputs, it is challenging to distinguish shape from appearance. For example, detailed structures such as clothing wrinkles can just as easily be painted as colors on a flat surface, leading to inaccurate surface geometry. To counteract this, we use normal estimators \cite{khirodkar2024sapiens} trained on large datasets to provide a signal to improve the geometry. We can take spatial derivatives of signed distance $d$ with respect to ${\bf X}_t$ to compute the surface normal of a point ${\bf X}_t$ in deformed space. We normalize the rendered and estimated surface normals and compute a normal loss as $\mathcal{L}_2$ error between them. Similar to prior work on neural surface reconstruction \cite{yariv2021volume}, we also compute an eikonal loss $\mathcal{L}_{\mathrm{eik}}$ to regularize the neural surface.
\begin{align}
\label{eq:normal-loss}
{\bf n} &= \mathrm{normalize}(\nabla{d(\mathcal{W}^{-}({\bf X}_t, t))}) \\
\mathcal{L}_{\bf n} &= \left\| \textbf{n} - \textbf{n}^* \right\|^2 =  2 - 2 \langle \textbf{n}, \textbf{n}^* \rangle\\
\mathcal{L}_{\bf eik} &= \left\|\mathrm{norm}(\nabla{d(\mathcal{W}^{-}({\bf X}_t, t))})-1\right\|
\end{align}

\noindent\textbf{Normals with numerical gradients.} Most prior work uses analytical gradients (e.g. auto-diff) to compute normals of signed distance fields. However, these are computed within an infinitesimally small neighborhood of ${\bf X}_t$ and suffer from noise in both the estimated backward warping fields and signed distances \cite{chetan2023accurate}. This leads to unstable optimization when dealing with deformable objects. To avoid this, we compute normals by numerical gradients \cite{li2023neuralangelo} with a fixed 1mm step size during optimization.

\noindent\textbf{Normals with eikonal filtering.} Although numerical normal computation works well on static scenes, it is more challenging in deformable scenes where the warping field's influence can cause $\|\textbf{X}_t + \delta - \textbf{X}_t\| $ to be very different from $\| \mathcal{W}^{-}(\mathbf{X}_t + \delta) - \mathcal{W}^{-}(\mathbf{X}_t)\|$. For example, the hand and waist might be close in deformed space but far in canonical space, causing exploding gradients due to a large change in signed distance gradient over a small neighborhood. To avoid this problem, we clip the normal direction to 0 after Eq. \ref{eq:normal-loss} whenever the gradient magnitude exceeds some threshold, in our case $\left\|\nabla d\right\|>10$.

\noindent{\bf Optical flow.} We use optical flow \cite{teed2020raft, yang2019volumetric} to learn the non-rigid deformation and relative camera transform between two frames. We compute 3D scene flow vectors by backward warping deformed points to canonical space, then forward-warping to another timestamp. We use the camera matrix to project 3D flow vectors into 2D, and compute $\mathcal{L}_2$ error between rendered flow $\mathbf f$ and estimated flow $\textbf{f}^*$, $L_{\textbf{f}} = \left\lVert\textbf{f}-\textbf{f}^*\right\rVert$. Here, $|t'-t|=\{1,2,4,8\}$:
\begin{align}
\textbf{f}_\text{3D}(\textbf{X}_t,t\to t') &= \mathcal{W}^+(\mathcal{W}^-(\textbf{X}_t,t),t')-\textbf{X}_t
\end{align}

\noindent{\bf Universal features.} Deep neural features are useful for registering pixels to a 3D model \cite{yang2021viser, lindenberger2021pixsfm}, while allowing better convergence at textureless regions or under deformation. Prior work relies on category-specific image features, but we find DINOv2 \cite{oquab2023dinov2}) to be a robust and universal feature descriptor that works well for clothing and accessories. We choose the small DINOv2 model with registers, as it produces fewer peaky feature artifacts \cite{darcet2023vision}. We obtain pixel-level features from DINOv2's patch descriptors by evaluating DINOv2 on an image pyramid, averaging features across pyramid levels, and reducing the dimension to 16 via PCA \cite{amir2021deep}. We compute feature loss as $\mathcal{L}_2$ error between rendered and estimated features, $\mathcal{L}_{\boldsymbol \phi} = \left\| {\boldsymbol \phi} - {\boldsymbol \phi}^* \right\|^2$.

\noindent\textbf{Zero-shot segmentation.} 
Inspired by shape-from-silhouette \cite{vlasic2008articulated}, we use image segmentation to carve out the 3D boundary of the target. We leverage the foundational 2D segmentation model SAM~\cite{kirillov2023segment} and its extension to tracking~\cite{yang2023track} to predict accurate silhouettes of humans with clothing and accessories. We pass different prompts according to different scenarios we aim to reconstruct, such as ``human wearing cloth'' and ``human holding an object''.
We compute silhouette loss as the $\mathcal{L}_2$ error between rendered and estimated silhouettes,
$\mathcal{L}_{\bf s} = \left\| {\bf s} - {\bf s}^* \right\|^2$.

\noindent\textbf{Losses.}
Our final loss is a weighted sum of reconstruction and regularization terms. Loss weights $\lambda$ are searched once and kept across all experiments.
\begin{align}
    \mathcal{L}_{\mathrm{rec}} &= \lambda_{\bf c}\mathcal{L}_{\bf c} + \lambda_{\bf f}\mathcal{L}_{\bf f} + \lambda_{\bf n}\mathcal{L}_{\bf n} + \lambda_{\boldsymbol \phi}\mathcal{L}_{\boldsymbol \phi} + \lambda_{\bf s}\mathcal{L}_{\bf s}\\
    \mathcal{L}_{\mathrm{reg}} &= \lambda_{\mathrm{eik}}\mathcal{L}_{\mathrm{eik}} + \lambda_{\mathrm{cyc}}\mathcal{L}_{\mathrm{cyc}} + \lambda_{\mathrm{sink}}\mathcal{L}_{\mathrm{sink}}+\lambda_{\mathrm{cl}}\mathcal{L}_{\mathrm{cl}}
\end{align}

\subsection{Refinement with 3D Gaussians}
\label{sec: refinement}
\noindent{\bf Representation.} Neural SDFs are ideal for extracting surfaces, but can be difficult to optimize as adding new geometry requires making global changes. In light of this, we introduce a refinement procedure that replaces the canonical shape representation with 3D Gaussians~\cite{kerbl20233dgs} while keeping the two-layer motion model as is. To render an image, we warp Gaussians forward from canonical space to time $t$ (Eq.~\ref{eq:lbs-fw}) and call the differentiable Gaussian rasterizer.

\noindent{\bf Initialization.} We use 40k Gaussians, each parameterized by 14 values, including its opacity, RGB color, center location, orientation, and axis-aligned scales. Gaussians are initialized on the surface of the neural SDF with isotropic scaling. To initialize the color of each Gaussian, we query the canonical color MLP (Eq.~\ref{eq:color}) at its center.

\noindent{\bf Optimization.} We update both the canonical 3D Gaussian parameters and the motion fields by minimizing
\begin{align}
    \mathcal{L}_{\mathrm{rec}}&=\lambda_{\bf c}\mathcal{L}_{\bf c} + \lambda_{\bf f}\mathcal{L}_{\bf f} + \lambda_{\bf s}\mathcal{L}_{\bf s}\\
    \mathcal{L}_{\mathrm{reg}} &= \lambda_{\mathrm{sink}}\mathcal{L}_{\mathrm{sink}}.
\end{align} 
Notably, the 3D cycle loss $\mathcal{L}_{\mathrm{cyc}}$ can be dropped since rasterization does not require computing backward warps. 

\section{Experiments}
We evaluate DressRecon's ability to reconstruct both 3D shape and appearance given challenging monocular videos. Video results are available on the supplementary webpage.

\subsection{Datasets}
\noindent\textbf{Dynamic clothing and accessories.} To evaluate DressRecon's ability to reconstruct dynamic clothing and objects, we select 14 sequences from DNA-Rendering \cite{cheng2023dna} with challenging cloth deformation and/or handheld objects (e.g. playing a cello, swinging a cloth, waving a brush). As DNA-Rendering does not provide ground-truth meshes, we compute pseudo-ground-truth 3D meshes by using all 48 available cameras to optimize a separate NeuS2 \cite{wang2023neus2} instance at each timestep. To overcome the limited viewpoint range of each individual camera, we assemble turntable monocular videos by rendering these per-frame NeuS2 instances along a smooth 360-degree camera trajectory.

\noindent\textbf{Avatars from casual videos.}
We also evaluate DressRecon's ability to recover high-fidelity human avatars from casual turntable videos. We evaluate our method on ActorsHQ \cite{isik2023humanrf} and select subsets of the first 4 sequences for evaluation, each about 200 frames. As ActorsHQ cameras have small fields of view and often do not cover the whole body, we colorize the provided ground-truth meshes and render turntable monocular videos with $360^\circ$ of camera rotation.

\subsection{Results}
\begin{figure}
\centering
\includegraphics[width=\linewidth, trim={0.1cm 0cm 7.0cm 0cm}, clip, angle=0]{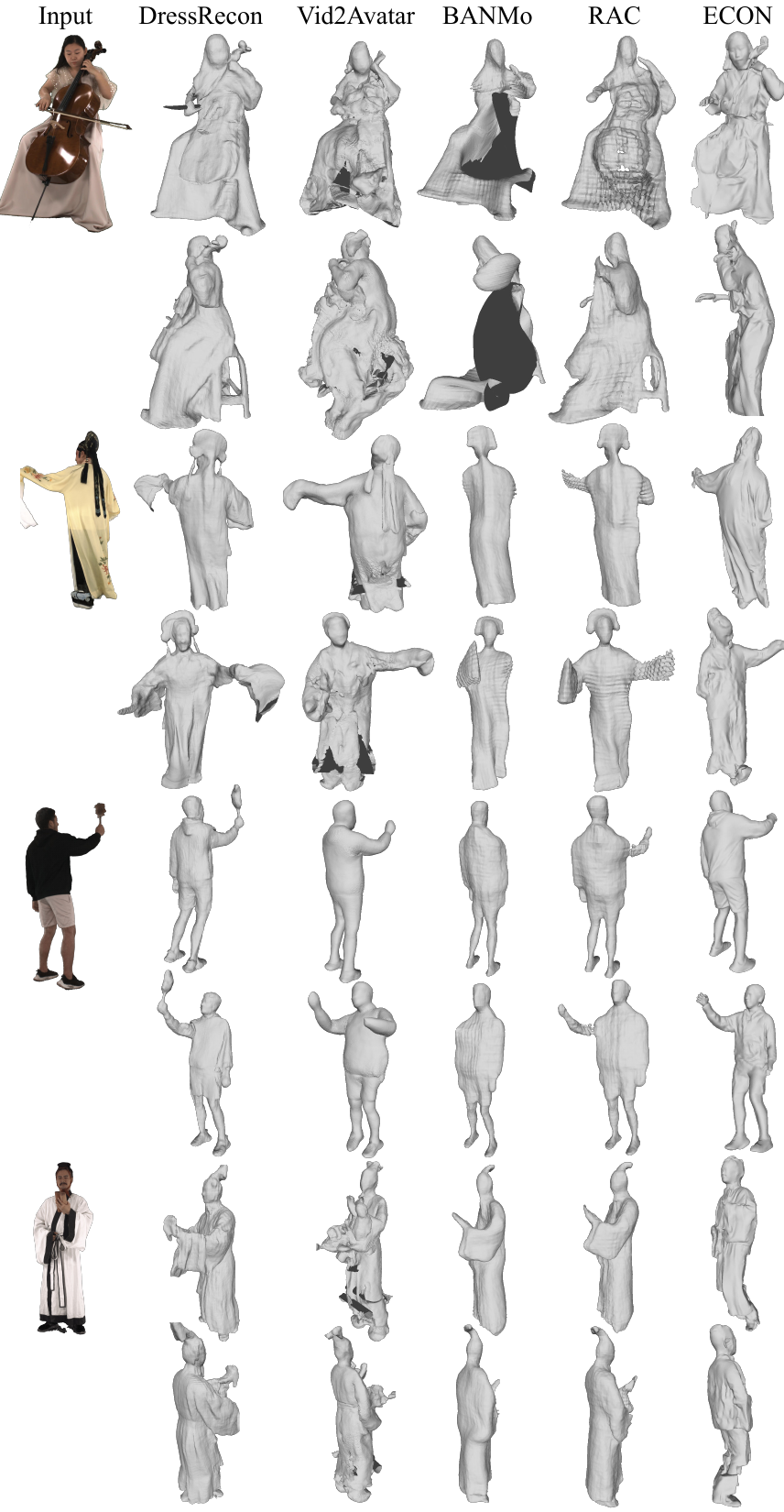}
\caption{
    \textbf{3D reconstruction results on DNA-Rendering.} We demonstrate DressRecon's ability to reconstruct challenging sequences with large cloth deformation. DressRecon's predictions align well with the image evidence, even in the presence of rapid clothing and object deformations. Vid2Avatar often outputs spurious shape artifacts and is unable to reconstruct challenging structures, such as the white cloth (row 2), brown brush (row 3), and detailed sleeves (row 4). BANMo and RAC produce hollow cellos on the first row, and tend to output over-smoothed surfaces for the other cases. ECON produces highly detailed textures, but it performs the worst numerically (Tab. \ref{tab:chamfer_dna_rendering}) as the outputs often have an incorrect overall shape (e.g. Row 1). We encourage readers to view the video results on the supplementary webpage.
}
\label{fig:dna_mesh_results}
\vspace{-20pt}
\end{figure}

\noindent\textbf{Reconstructing dynamic clothing and accessories.}
Tab. \ref{tab:chamfer_dna_rendering} reports the 3D chamfer distance (cm, $\downarrow$) for reconstructing dynamic clothing and handheld objects, evaluated across 14 DNA-Rendering sequences. We compare with Vid2Avatar \cite{guo2023vid2avatar}, BANMo \cite{yang2022banmo}, RAC \cite{yang2023rac}, and ECON \cite{xiu2023econ}, and show qualitative results in Fig. \ref{fig:dna_mesh_results}. The \href{https://jefftan969.github.io/dressrecon/}{project page} contains corresponding video results. DressRecon reconstructs finer details and more accurate body shape than prior art, and is able to handle challenging scenarios such as the tip of the cello (image 1), the hair tassels (image 2), and the detailed cloth wrinkles on the martial arts uniform (image 4).

\begin{figure}
\centering
\includegraphics[width=0.95\linewidth, trim={0cm 13.5cm 9.5cm 0cm}, clip, angle=0]{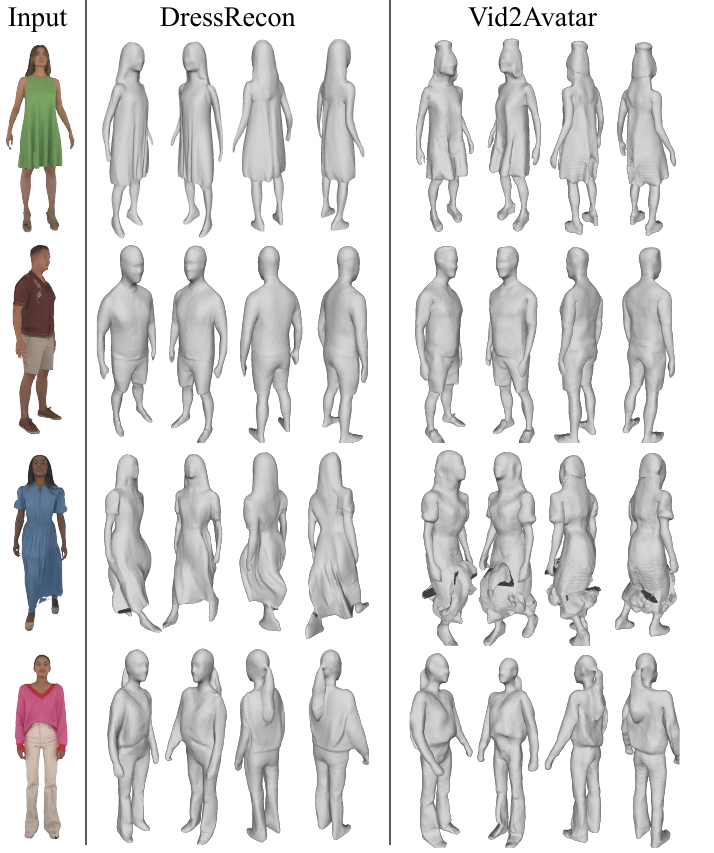}
\caption{
    \textbf{3D reconstruction results on ActorsHQ.} DressRecon is on par with Vid2Avatar for standard clothing (Rows 2 and 4), and higher fidelity than Vid2Avatar for loose clothing (Rows 1 and 3). Vid2Avatar's reconstructed skirts often contain shape artifacts. We attribute DressRecon's improved performance to its flexible shape and deformation representation, which is capable of representing non-standard geometry and deformation.
}
\label{fig:actorshq_mesh_results}
\vspace{-10pt}
\end{figure}

\noindent\textbf{Reconstructing avatars from casual videos.}
Tab. \ref{tab:chamfer_actorshq} reports 3D chamfer distance (cm, $\downarrow$) and F-score at $\{1,2,5\}$-cm thresholds for recovering avatars from turntable videos, evaluated across 4 ActorsHQ sequences. We compare with Vid2Avatar \cite{guo2023vid2avatar} and show qualitative results in Fig. \ref{fig:actorshq_mesh_results}. DressRecon performs on par with Vid2Avatar in tight clothing scenarios, and reconstructs higher-fidelity geometry on sequences with challenging clothing such as dresses.

\begin{figure*}
\centering
\includegraphics[width=0.8\linewidth, trim={0cm 13.5cm 0cm 0cm}, clip, angle=0]{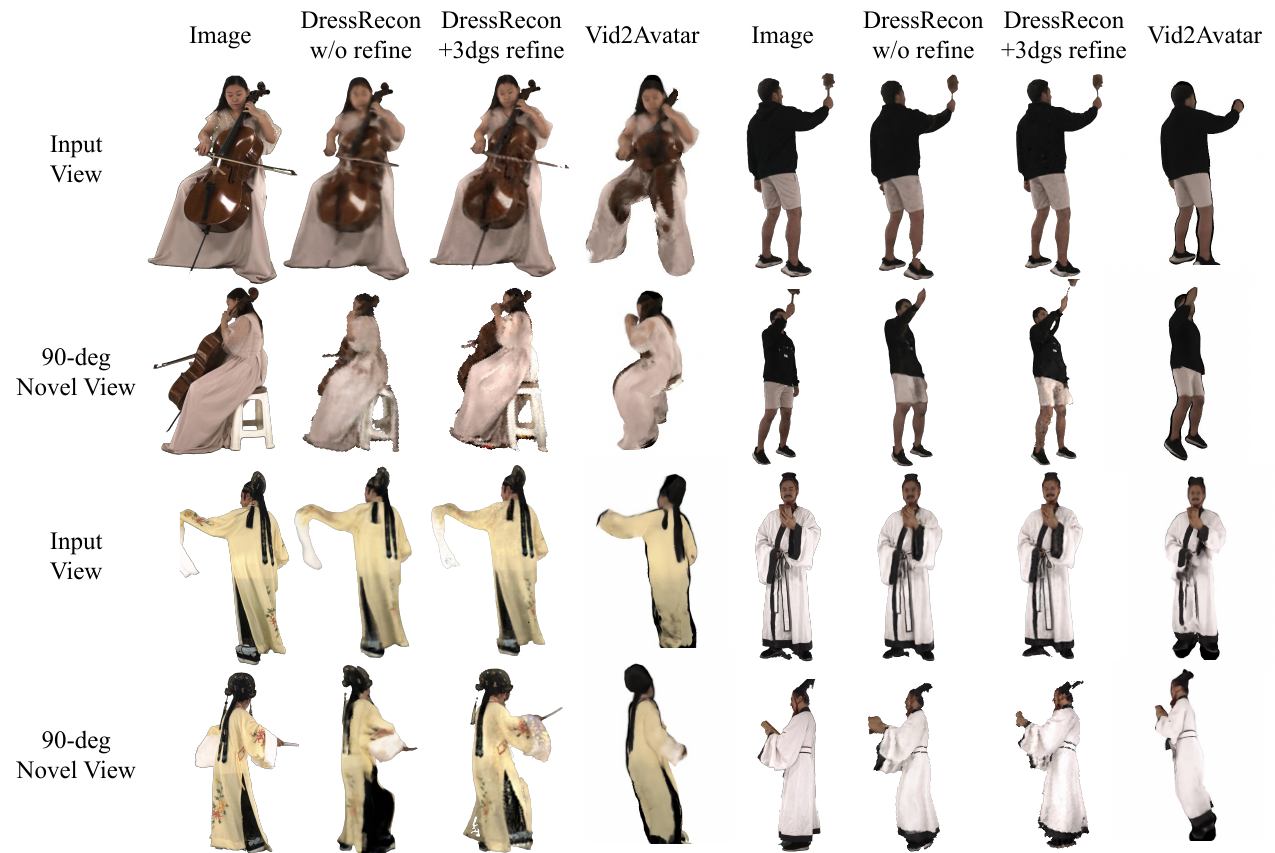}
\caption{
    \textbf{RGB rendering results on DNA-Rendering.} For each sequence, we show DressRecon's and Vid2Avatar's renderings at both the input view and a 90-degree novel view. DressRecon's renderings are shown with and without 3D Gaussian refinement. We find (similar to Tab. \ref{tab:psnr_dna_rendering}) that refinement significantly improves the textures, especially the flowers on the yellow dancer's sleeve. Vid2Avatar's renderings are less detailed, and fail to accurately depict structures that substantially deviate from the body, such as the cello and white stool.
}
\label{fig:dna_rgb_results}
\vspace{-10pt}
\end{figure*}

\noindent\textbf{Rendering dynamic clothing and accessories.}
Tab. \ref{tab:psnr_dna_rendering} reports the RGB PSNR ($\uparrow)$, SSIM ($\uparrow$), LPIPS ($\downarrow$), and mask IoU ($\uparrow$) on test views by holding out every 8-th training view. We compare against Vid2Avatar \cite{guo2023vid2avatar}, BANMo \cite{yang2022banmo}, and RAC \cite{yang2023rac}, and show qualitative results in Fig. \ref{fig:dna_rgb_results}. The \href{https://jefftan969.github.io/dressrecon/}{project page} contains extensive video results. DressRecon produces more accurate renderings than prior art.

\begin{figure}[t!]
\begin{center}
\includegraphics[width=0.5\linewidth, trim={5.1in 5.85in 1.1in 3.5in}, clip]{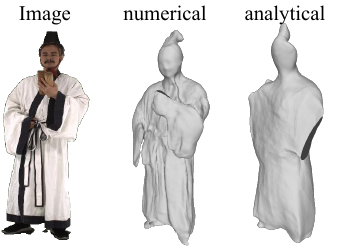}
\end{center}
\vspace{-10pt}
\caption{
    \textbf{Qualitative ablation of numerical normals}. We show the difference between optimizing with numerical and analytical normals. Using analytical normals causes training to be unstable, resulting in a flat shape with no surface detail. The quality of surface details is reduced when normal loss is disabled (Tab. \ref{tab:ablation_dna_mesh}).
}
\label{fig:ablation_normals}
\vspace{-10pt}
\end{figure}

\begin{table}
    \caption{
        \textbf{3D reconstruction metrics on DNA-Rendering sequences.} We evaluate 3D chamfer distance (cm, $\downarrow$) on fourteen DNA-Rendering sequences with challenging clothing deformation or handheld objects. DressRecon outperforms all baselines, and is the \textbf{best} or \underline{second-best} method on all sequences.
    }
    \small
    \centering
    \setlength{\tabcolsep}{4pt}
    \begin{tabular}{lccccc}
    \toprule
    Sequence & DressRecon & Vid2Avatar & BANMo & RAC & ECON\\
    & (Ours) & \cite{guo2023vid2avatar} & \cite{yang2022banmo} & \cite{yang2023rac} & \cite{xiu2023econ}\\
    \midrule
    0008\_01 & \underline{7.300} & \textbf{6.786} & 7.768 & 8.112 & 9.420\\
    0047\_01 & \underline{8.542} & 11.216 & \textbf{7.808} & 9.106 & 17.102\\
    0047\_12 & \textbf{6.064} & 9.334 & 6.914 & \underline{6.618} & 7.541\\
    0102\_02 & \underline{5.421} & 7.812 & \textbf{5.131} & 7.278 & 10.181\\
    0113\_06 & \textbf{6.872} & 8.517 & \underline{8.362} & 8.391 & 11.282\\
    0121\_02 & \textbf{5.520} & \underline{6.478} & 7.453 & 6.926 & 9.334\\
    0123\_02 & \textbf{6.725} & 8.343 & \underline{7.418} & 9.683 & 12.108\\
    0128\_04 & \textbf{7.803} & \underline{8.184} & 8.913 & 10.005 & 11.569\\
    0133\_07 & \textbf{6.194} & \underline{6.314} & 7.017 & 7.449 & 7.320\\
    0152\_01 & \underline{6.437} & 7.465 & 6.815 & \textbf{6.170} & 8.735\\
    0166\_04 & \textbf{4.356} & \underline{4.608} & 5.969 & 6.286 & 6.562\\
    0188\_02 & \textbf{5.403} & 5.887 & 6.341 & \underline{5.829} & 9.026\\
    0206\_04 & \textbf{7.555} & \underline{8.392} & 8.404 & 9.644 & 9.987\\
    0239\_01 & \underline{5.559} & \textbf{5.503} & 6.503 & 7.831 & 8.026\\
    \midrule
    Average & \textbf{6.411} & 7.489 & \underline{7.201} & 7.809 & 9.871\\
    \bottomrule
    \label{tab:chamfer_dna_rendering}
\end{tabular}
\vspace{-10pt}
\end{table}

\begin{table}
    \caption{
        \textbf{Rendering metrics on DNA-Rendering sequences.} We evaluate RGB PSNR ($\uparrow$), SSIM ($\uparrow$), LPIPS ($\downarrow$), and mask IoU ($\uparrow$), averaged across fourteen DNA-Rendering sequences with challenging clothing deformation or handheld objects. DressRecon outperforms all baselines, particularly when 3D Gaussian refinement is used to improve the rendering quality.
    }
    \small
    \centering
    \setlength{\tabcolsep}{4pt}
    \begin{tabular}{lcccc}
    \toprule
    Method & PSNR & SSIM & LPIPS & Mask IoU\\
    \midrule
    DressRecon & \underline{22.03} & \underline{0.9375} & 
\underline{0.1059} & \underline{0.9544}\\
    w/ 3DGS refinement & \textbf{22.27} & \textbf{0.9506} & \textbf{0.0860} & \textbf{0.9786}\\
    \midrule
    Vid2Avatar \cite{guo2023vid2avatar} & 19.61 & 0.8948 & 0.1167 & 0.7931\\
    BANMo \cite{yang2022banmo} & 19.78 & 0.9341 & 0.1257 & 0.8988\\
    RAC \cite{yang2023rac} & 19.89 & 0.9351 & 0.1157 & 0.9044\\
    \bottomrule
\label{tab:psnr_dna_rendering}
\end{tabular}
\vspace{-10pt}
\end{table}

\begin{table}
    \caption{
        \textbf{3D reconstruction metrics on ActorsHQ sequences.} We evaluate 3D chamfer distance (cm, $\downarrow$) and F-score at $\{1,2,5\}$-cm thresholds (\%, $\uparrow$) on the four ActorsHQ sequences shown in Fig. \ref{fig:actorshq_mesh_results}. DressRecon outperforms Vid2Avatar on most sequences.
    }
    \footnotesize
    \centering
    {%
    \setlength{\tabcolsep}{4pt}
    \begin{tabular}{l|cccc|cccc}
    \toprule
    & \multicolumn{4}{c}{DressRecon} & \multicolumn{4}{c}{Vid2Avatar}\\
    Sequence & CD & F@5 & F@2 & F@1 & CD & F@5 & F@2 & F@1\\
    \midrule
    a1s1 & 3.212 & 94.74 & \textbf{72.46} & \textbf{48.99} & \textbf{3.204} & \textbf{98.22} & 69.69 & 39.25 \\
    a2s1 & \textbf{1.838} & \textbf{99.96} & \textbf{92.72} & \textbf{62.14} & 2.891 & 97.59 & 79.35 & 40.65\\
    a3s1 & \textbf{2.647} & \textbf{97.15} & \textbf{79.91} & \textbf{51.38} & 4.376 & 90.93 & 56.83 & 30.33\\
    a4s1 & 2.247 & \textbf{98.88} & 85.02 & 57.00 & \textbf{2.039} & 98.82 & \textbf{89.26} & \textbf{63.84}\\
    \midrule
    Average & \textbf{2.486} & \textbf{97.68} & \textbf{82.53} & \textbf{54.88} & 3.128 & 96.39 & 73.78 & 43.52
    \\\bottomrule
    \end{tabular}
    }
\label{tab:chamfer_actorshq}
\vspace{-10pt}
\end{table}

\subsection{Diagnostics}

\noindent\textbf{3D Gaussian refinement.}
In Tab. \ref{tab:ablation_dna_rgb}, we show results from optimizing a neural implicit model from scratch, a 3D Gaussian model from scratch, and a 3D Gaussian model initialized from a neural implicit model. The same computational budget is allocated to all three experiments. The highest rendering quality is achieved with neural implicit optimization followed by 3D Gaussian refinement. This suggests that the neural implicit model helps produce a good initialization of shape and deformation, making it easier for 3DGS to converge to better local optima.

\noindent\textbf{Choice of deformation model.}
In Tab. \ref{tab:ablation_dna_mesh}, we swap our hierarchical two-layer deformation model with several alternatives in the literature. Swapping to a skeleton+dense warping field \cite{jiang2022selfrecon, weng2022humannerf}, skeleton alone \cite{yang2023rac}, or bag-of-bones alone \cite{yang2022banmo} reduces the geometry quality. Alternative deformation models are also less interpretable, as skeleton-only and bag-of-bones do not separate body and clothing motion.

\noindent\textbf{Choice of image-based priors.} In Tab. \ref{tab:ablation_dna_mesh}, we run the optimization routine and remove one of the image-based priors each time. Without mask loss, the surface geometry has an incorrect overall structure. Without normal loss, the reconstructed surface has lower detail. Without flow loss, the shape is less sensible and camera optimization is less stable.

\noindent\textbf{Choice of normal supervision.}
In Fig. \ref{fig:ablation_normals}, we show the benefit of using normal loss with numerical gradients. With analytical gradients, shape optimization becomes unstable.

\begin{table}
    \caption{
        \textbf{Ablation study for 3D reconstruction.} We ablate the importance of motion field representation and choice of image-based priors, by evaluating 3D chamfer distance (cm, $\downarrow$) and F-score at $\{1,2,5\}$-cm thresholds (\%, $\uparrow$) on 14 DNA-Rendering sequences. DressRecon performs worse after switching motion representations (skeleton-only \cite{guo2023vid2avatar}, bag-of-bones \cite{yang2022banmo}, skeleton+dense \cite{jiang2022selfrecon}) and after removing any image-based prior.
    }
    \small
    \centering
    {%
    \setlength{\tabcolsep}{4pt}
    \begin{tabular}{r|cccc}
    \toprule
    Sequence & CD & F@5 & F@2 & F@1\\
    \midrule
    DressRecon & \textbf{6.411} & \textbf{81.16} & \textbf{47.66} & \textbf{25.62}\\
    \midrule
    skeleton-only & 7.340 & 78.10 & 44.55 & 22.67\\
    bag-of-bones & 6.942 & 80.19 & 47.21 & 25.49\\
    skeleton+dense & 7.526 & 76.33 & 41.58 & 21.91\\
    \midrule
    w/o mask & 10.647 & 65.73 & 33.61 & 17.42\\
    w/o normal & 7.206 & 77.70 & 43.46 & 23.53\\
    w/o flow & 7.094 & 79.03 & 45.72 & 24.41\\
    w/o pose & 6.938 & 78.87 & 46.21 & 25.11\\
    w/o feat & 6.829 & 79.25 & 46.75 & 25.34
    \\\bottomrule
    \end{tabular}
    }
\label{tab:ablation_dna_mesh}
\vspace{-10pt}
\end{table}

\begin{table}
    \caption{
        \textbf{Ablation study for Gaussian refinement.} We ablate the impact of 3D Gaussian refinement, by evaluating RGB PSNR ($\uparrow$), SSIM ($\uparrow$), LPIPS ($\downarrow$), and mask IoU ($\uparrow$) on 14 DNA-Rendering sequences. We perform experiments where only an implicit SDF is optimized, where 3D Gaussians are optimized without initializing from an SDF, and where a neural SDF is used to initialize 3D Gaussians. The best rendering quality is obtained by initializing 3D Gaussians from an SDF.
    }
    \small
    \centering
    {%
    \setlength{\tabcolsep}{4pt}
    \begin{tabular}{r|cccc}
    \toprule
    Sequence & PSNR & SSIM & LPIPS & Mask IoU\\
    \midrule
    Implicit-only & 22.03 & 0.9375 & 0.1059 & 0.9544\\
    3DGS-only & 21.31 & 0.9455 & 0.0939 & 0.9737\\
    Implicit$\rightarrow$3DGS & \textbf{22.27} & \textbf{0.9506} & \textbf{0.0860} & \textbf{0.9786}\\
    \bottomrule
    \end{tabular}
    }
\label{tab:ablation_dna_rgb}
\vspace{-10pt}
\end{table}

\section{Discussion}
We present DressRecon, which reconstructs humans with loose clothing and accessory objects from monocular videos. DressRecon uses hierarchical bag-of-bones deformation to model clothing and body deformation separately, and leverages off-the-shelf priors such as masks and surface normals to make optimization more tractable. To improve the rendering quality, we introduce a refinement stage that converts the implicit neural body into 3D Gaussians. 

\noindent\textbf{Limitations.}
DressRecon requires sufficient view coverage to reconstruct a complete human, and cannot hallucinate unobserved body parts. It also has no understanding of cloth deformation physics. As a result, clothing may deform unnaturally if we reanimate with novel body motion. We leave reanimating human-cloth and human-object interactions as future work. Moreover, specifying inaccurate segmentation, e.g. by passing the wrong prompt to SAM \cite{kirillov2023segment}, could result in failure to reconstruct some details.

{
    \small
    \bibliographystyle{ieeenat_fullname}
    \bibliography{main}
}
\clearpage
\setcounter{page}{1}
\maketitlesupplementary

\section{Video Results}
Please see the attached webpage for video results.

\section{Implementation Details}

\subsection{Consistent 4D Neural Fields}
\noindent\textbf{Signed distance fields.} We initialize canonical signed distance fields as a sphere with radius 0.1m. Following standard practice, we apply positional encodings to all 3D points ($L_{xyz}=10$) and timestamps ($L_t=6$) before passing into MLPs. The appearance code $\boldsymbol{\omega}_t$ has 32 channels.

After $\bf{MLP}_{\mathrm{SDF}}$ computes the signed distance $d$ at a 3D point, we convert the signed distance to a volumetric density $\sigma\in[0,1]$ for volume rendering. Similar to VolSDF \cite{yariv2021volume}, this is done using the cumulative Laplace distribution $\sigma=\Gamma_\beta(d)$, where $\beta$ is a global learnable scalar parameter that controls the solidness of the object, approaching zero for solid objects. This representation allows us to extract a mesh as the zero level-set of the SDF.

\noindent\textbf{Cycle consistency regularization.} Given a forward warping field $\mathcal{W}^+(t):\mathbf{X}\to\mathbf{X}_t$ and a backward warping field $\mathcal{W}^-(t):\mathbf{X}_t\to\mathbf{X}$, we introduce a cycle consistency term, similar to NSFF \cite{li2021neural}. A sampled 3D point in camera coordinates should return to its original location after passing through a backward and forward warping:

\begin{equation}
\mathcal{L}_{\mathrm{cyc}}=\sum_{\mathbf{X}_t}\|\mathcal{W}^+(\mathcal{W}^-(\mathbf{X}_t,t),t)-\mathbf{X}_t\|_2^2
\end{equation}

\subsection{Hierarchical Gaussian Motion Fields}

\noindent\textbf{Bag-of-bones skinning deformation.} Our motion model uses the motion of $B$ bones (defined as 3D Gaussians, typically $B=25$) to drive the motion of canonical geometry. Given 3D Gaussians, we compute dense 3D motion fields by blending the $\mathbf{SE}(3)$ transformations of canonical Gaussians with skinning weights $\mathbf{W}$:

\begin{align}
{\bf X}_t &= \mathcal{W}^{+}({\bf X}, t) = \left(\sum_{b=1}^B {\bf W}^{+,b}{\bf G}^{b}_t\left({\bf G}^b\right)^{-1}\right){\bf X}\label{eq:lbs-fw} \\
{\bf X} &= \mathcal{W}^{-}({\bf X}_t, t)=\left(\sum_{b=1}^B {\bf W}^{-,b}_t{\bf G}^b\left({\bf G}^b_t\right)^{-1}\right){\bf X}_t\label{eq:lbs-bw}
\end{align}

\noindent where $\mathbf{G}^+_t$ are forward warps from canonical to time $t$ Gaussians, $\mathbf{G}^-_t$ are backward warps from time $t$ to canonical Gaussians, and $\mathbf{W}^+$ are forward skinning weights.

Similar to SCANimate \cite{saito2021scanimate} and LASR \cite{yang2021lasr}, we define a forward skinning weight function $\mathcal S^+:\mathbf{X}\to\mathbb{R}^B$ which computes the normalized influence of each Gaussian bone on a canonical 3D point. At a coarse level, skinning weights are defined as the Mahalanobis distance from $\mathbf{X}$ to the canonical Gaussians:
\begin{equation}\label{eq:lbs_coarse_fw}
\mathbf{W}^+_\sigma=(\mathbf{X}-\boldsymbol{\mu})^\top\mathbf{Q}(\mathbf{X}-\boldsymbol{\mu}),
\end{equation}
\noindent where $\boldsymbol{\mu}\in\mathbf{R}^{B\times 3}$ are canonical bone centers, $\mathbf{Q}=\mathbf{V}^\top \mathbf{\Lambda} \mathbf{V}$ are canonical bone precision matrices, $\mathbf{V}\in\mathbf{R}^{B\times\mathbf{SO}(3)}$ are canonical bone orientations, and $\mathbf{\Lambda}^{B\times 3\times 3}$ are time-invariant axis-aligned diagonal scale matrices.

In addition to a coarse component, we find it helpful to use delta skinning weights to model fine geometry. Delta skinning weights are computed by a coordinate MLP:
\begin{equation}\label{lbs_fine_fw}
\mathbf{W}^+_\Delta=\mathbf{MLP}_{\Delta,+}(\mathbf{X},t)\in\mathbb{R}^B
\end{equation}
The final skinning function is a normalized sum of coarse and fine components:
\begin{equation}\label{lbs_coarse_fine_fw}
\mathbf{W}^+=\mathcal{S}^+(\mathbf{X},t)=\text{softmax}(-\mathbf{W}^+_\sigma-\mathbf{W}^+_\Delta),
\end{equation}
where the negative sign ensures that faraway Gaussian bones (which have a larger Mahalanobis distance) are assigned a lower skinning weight after softmax.

Backward skinning weights are computed analogously with the time $t$ Gaussians, which have center $\boldsymbol{\mu}_t$, orientation $\mathbf{V}_t$, and time-invariant scale $\mathbf{\Lambda}$. We also need the transformation $\mathbf{G}^-_t$ from each time $t$ Gaussian to the canonical Gaussian, as well as the backward skinning $\mathbf{MLP}^-_\Delta$.

\begin{table*}
    \caption{\textbf{Summary of losses and loss weights.} Our final loss is a weighted sum of reconstruction terms (color, optical flow, normal, feature, and segmentation) and regularization terms (eikonal, cycle-consistency, gaussian consistency, camera prior, and joint prior).}
    \centering
    \begin{tabular}{ccl}
	\toprule
        Loss &Weight &Description\\
        \midrule
        $\mathcal{L}_\mathbf{c}$ & $\lambda_\mathbf{c}=0.1$ & L2 loss, rendered RGB vs. the input image\\
        $\mathcal{L}_\mathbf{f}$ & $\lambda_\mathbf{f}=0.5$ & L2 loss, rendered 2D flow vs. computed flow from VCNPlus \cite{yang2019volumetric}\\
        $\mathcal{L}_\mathbf{n}$ & $\lambda_\mathbf{n}=0.03$ & L2 loss, rendered normals vs. computed normals from Sapiens \cite{khirodkar2024sapiens}\\
        $\mathcal{L}_\mathbf{\phi}$ & $\lambda_\mathbf{\phi}=0.01$ & L2 loss, rendered features vs. computed features from DINOv2 \cite{oquab2023dinov2}\\
        $\mathcal{L}_\mathbf{s}$ & $\lambda_\mathbf{s}=0.1$ & L2 loss, rendered masks vs. computed masks from SAM \cite{kirillov2023segment}\\
        \midrule
        $\mathcal{L}_\text{eik}$ & $\lambda_\text{eik}=0.01$ & Encourage numerical gradients of canonical SDF to have unit norm\\
        $\mathcal{L}_\text{cyc}$ & $\lambda_\text{cyc}=0.05$ & Encourage backward and forward warping fields to be inverses\\
        $\mathcal{L}_\text{gauss}$ & $\lambda_\text{gauss}=0.2$ & Sinkhorn divergence between canonical 3D Gaussians and SDF\\
        $\mathcal{L}_\text{cloth}$ & $\lambda_\text{cloth}=0.1$ & Minimize the magnitude of clothing deformation\\
        \bottomrule
    \end{tabular}
    \label{tab:loss_summary}
 \end{table*}

\subsection{Optimization}

\noindent\textbf{Sampling.} Due to the expensive per-ray computation in volume rendering, optimization with batch gradient descent is challenging. As a result, previous methods randomly sample entire images~\cite{yang2023rac} to compute the reconstruction terms, leading to small batch sizes (typically 16 images per batch) and noisy gradients. We implement an efficient data-loading pipeline with memory-mapping that allows per-pixel measurements (e.g., RGB, flow, features) to load directly from disk without accessing the full image. This allows loading pixels from significantly more images in a single batch (e.g. 256 images on a GPU).

\noindent\textbf{Hyperparameters.} We use the Adam optimizer with learning rate 0.0005. We use 48k iterations of optimization for all experiments. On a single RTX 4090 GPU, it takes about 8 hours to optimize the neural implicit body model and 15 seconds to render each frame. 3D Gaussian refinement is performed for another 48k iterations of optimization, taking about 8 hours to optimize and 0.1 seconds to render each frame. Our loss weights are described in Tab. \ref{tab:loss_summary}. At each iteration, we sample 72 images and take 16 pixel samples per image. For training efficiency, input images are cropped to a tight bounding box around the object and resized to 256x256. To prevent floater artifacts from appearing outside the tight crop, 90\% of pixel samples are taken from the tight bounding box and 10\% of pixel samples are taken from the full un-cropped image.

\end{document}